\pdfoutput=1
\documentclass[11pt]{article}

\usepackage{acl}

\usepackage{times}
\usepackage{latexsym}
\usepackage{subfig}
\usepackage[T1]{fontenc}
\usepackage{changepage,threeparttable} % for wide tables
\usepackage{booktabs, multirow} % for borders and merged ranges
\usepackage{makecell}
\usepackage{tablefootnote}
\usepackage[utf8]{inputenc}
\usepackage{amssymb}

\usepackage{microtype}
\usepackage{xspace}
\usepackage{arydshln} 
\usepackage{hyperref}

\usepackage[most]{tcolorbox}
\usepackage{enumerate}
\newenvironment{itemize*}%
 {\leftmargini=10pt\begin{itemize}%
  \setlength{\itemsep}{0pt}%
  \setlength{\parskip}{0pt}%
  }%
 {\end{itemize}}
\newenvironment{enumerate*}%
 {\begin{enumerate}%
  \setlength{\itemsep}{0pt}%
  \setlength{\parskip}{0pt}}%
 {\end{enumerate}}

\definecolor{myblue}{rgb}{0.9, 0.1, 0.94}
\definecolor{myyellow}{rgb}{0.98, 0.94, 0.75}
\definecolor{mygreen}{rgb}{0.68, 0.9, 0.6}
\definecolor{myorange}{rgb}{1.0, 0.49, 0.0}

\newcommand{\toolname}{\textsc{DataLab}\xspace}

\usepackage{listings}
\usepackage{color}

\definecolor{dkgreen}{rgb}{0,0.6,0}
\definecolor{gray}{rgb}{0.5,0.5,0.5}
\definecolor{mauve}{rgb}{0.58,0,0.82}

\definecolor{codepink}{rgb}{0.98, 0.25, 0.3}
\definecolor{codeblue}{rgb}{0.00, 0.53, 0.99}
\definecolor{codegreen}{rgb}{0.38,0.1,0.82}

\title{\toolname: A Platform for Data Analysis and Intervention}

 \author{Yang Xiao$^\clubsuit$\thanks{\ \  Work done during a remote research collaboration with CMU} , \;\; Jinlan Fu$^\bigstar$, \;\; Weizhe Yuan$^\spadesuit$, \;\; Vijay Viswanathan$^\spadesuit$,   \\ \bf Zhoumianze Liu$^\clubsuit$, \;\;  Yixin Liu$^\blacktriangle$, \;\; Graham Neubig$^{\spadesuit\dag}$, \;\; Pengfei Liu$^{\spadesuit}$ \thanks{\ \ Corresponding author} \\ 
 \\
  $^\spadesuit$Carnegie Mellon University,  $^\clubsuit$Fudan University, $^\bigstar$National University of Singapore, \\ $^\blacktriangle$Yale University, $^\dag$Inspired Cognition \\ 
  }

\begin{document}
\maketitle
\begin{abstract}

Despite data's crucial role in machine learning, most existing tools and research tend to focus on systems on top of existing data rather than how to interpret and manipulate data.
In this paper, we propose \toolname, a unified data-oriented platform
that not only allows users to interactively analyze the characteristics of data, but also provides a standardized interface for different data processing operations. Additionally, in view of the ongoing proliferation of datasets, \toolname has features for dataset recommendation and global vision analysis that help researchers form a better view of the data ecosystem.
So far, \toolname covers 1,715 datasets 
and 3,583 of its transformed version (e.g., hyponyms replacement ),
where 728 
datasets support various analyses (e.g., with respect to gender bias) with the help of 
140M samples annotated by 318 feature functions. 
\toolname is under active development and will be supported going forward. We have released a 
\textit{web platform},\footnote{\url{http://datalab.nlpedia.ai/}} 
web \href{https://app.swaggerhub.com/apis-docs/XiaoYang66/Datalab/1.0.0}{API},
Python SDK,\footnote{\url{https://github.com/ExpressAI/DataLab}},
\textit{PyPI}\footnote{\url{https://pypi.org/project/datalabs/}} published package and online documentation,\footnote{\url{https://expressai.github.io/DataLab/}} which hopefully, can meet the diverse needs of researchers. 
%calculated

\end{abstract}

\section{Introduction}

Datasets power modern natural language processing (NLP) systems, playing an essential role in model training, evaluation, and deployment~\cite{paullada2021data}.
Furthermore, methods to process data and understand have been subject to much research, including on topics such as data augmentation \cite{fadaee-etal-2017-data,feng_2021}, adversarial evaluation \cite{jia-liang-2017-adversarial,marco_2021}, bias analysis \cite{zhao_2018,blodgett-etal-2020-language}, and prompt-based learning \cite{liu_2021}.
Despite the critical role of data in NLP, the majority of open-source tooling regarding NLP has focused on methods to \emph{build models given data}, rather than to \emph{analyze and intervene upon the data itself}. In this paper, we present \toolname, a unified platform that allows NLP researchers to perform a number of data-related tasks in an efficient and easy-to-use manner:

\begin{figure}
    \centering
    \includegraphics[width=0.85\linewidth]{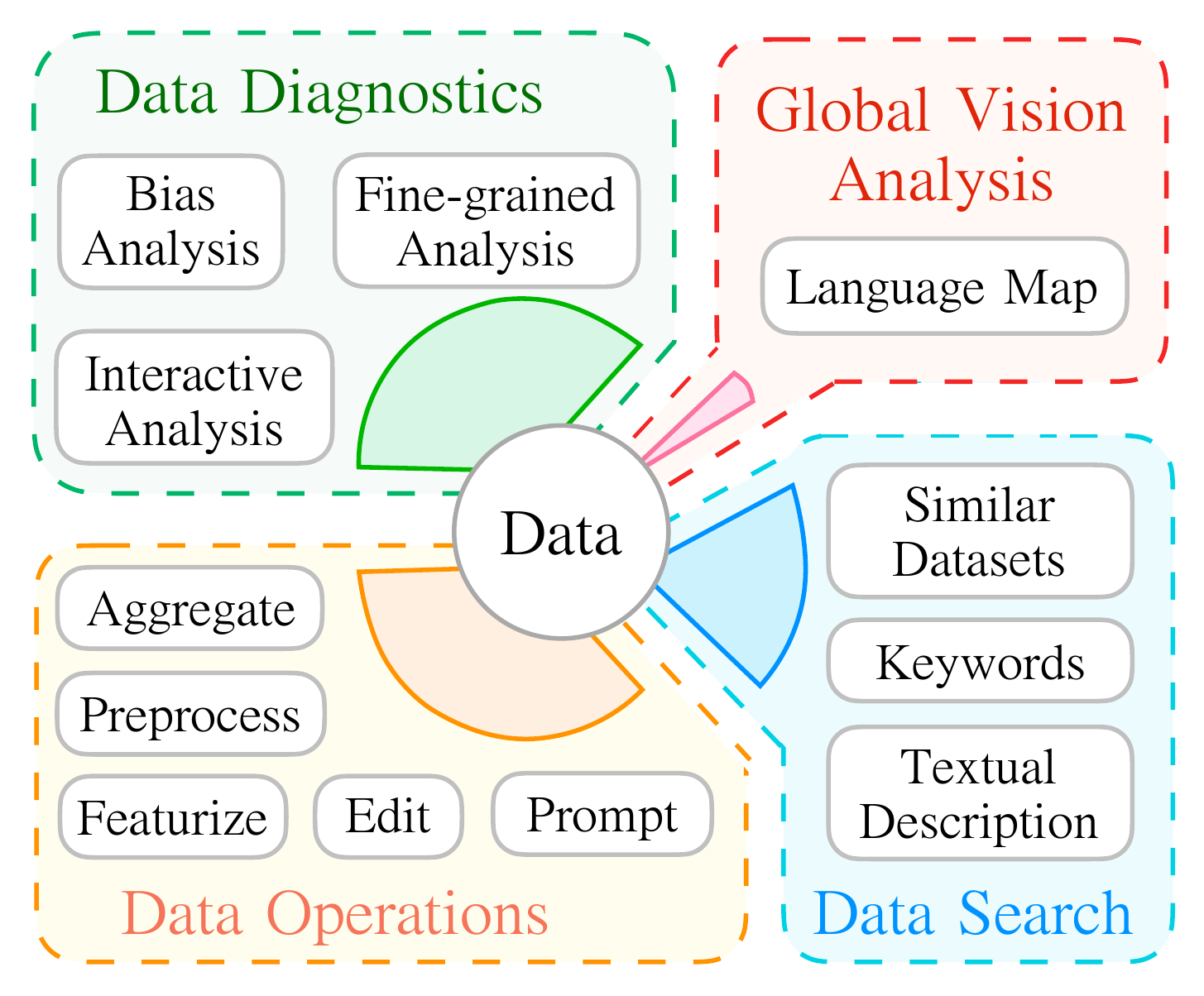}
    \caption{Overview of \toolname functionality}
    \label{fig:overview}
\end{figure}

(1) \textbf{Data Diagnostics}: While a significant amount of research has focused on interpreting the outputs of machine learning systems \citep{lipton2018mythos,belinkov-glass-2019-analysis}, data deserves deeper understanding as a first-class citizen of the machine learning ecosystem. \toolname allows for analysis and understanding of data to uncover undesirable traits such as hate speech, gender bias, or label imbalance (as shown in Fig.\ref{fig:overview} and \autoref{sec:diagnostics}).

\noindent (2) \textbf{Operation Standardization}:
There are a number of well-designed packages for data-oriented operations such as preprocessing \citep{loper2002nltk,manning2014stanford,TakuKudo} or editing \citep{marco_2021,dhole2021nl}. In practice, however, the diversity of requirements makes it necessary for users to install a variety of packages that use different data processing interfaces.
This (a) reduces the efficiency of development, (b) can confuse users (e.g., not knowing what preprocessing methods are appropriate for a given dataset?), and (c) is detrimental for reproducibility \citep{marie-etal-2021-scientific}.
\toolname provides and standardizes a large number of data processing operations, assigning each operation a unique identifier, to mitigate these issues  (\autoref{sec:operations}).

%commonly it is difficult to find accurate data preprocessing or evaluation tools from the description of the paper.
\noindent (3) \textbf{Data Search}
% In addition to a comprehensive understanding of datasets and standardized operations, an
An important question in practice is which datasets to use in a given scenario, given the huge proliferation of datasets in recent years.%
\footnote{According to \href{https://paperswithcode.com/}{Papers With Code}, the number of AI-related academic datasets has doubled in the past two years.}
\toolname provides a semantic dataset search tool to help identify appropriate datasets (\autoref{sec:search}).

\noindent (4) \textbf{Global Analysis}
% Data is not only the basis of model design training, but also the primary medium for benchmarking and evaluation. In addition to the analysis of the data itself, we can also mine the information reflected through the data from the global view （bird's-eye view）.
Beyond individual datasets, analyzing the entire ecosystem of existing datasets as a whole can yield insights. From a birds-eye view, we can get a clearer picture: \textit{where we are} and \textit{where efforts should be focused} to avoid systemic inequalities \cite{blodgett-etal-2020-language,blasi2021systematic}.
\toolname provides tools to perform similar global analyses over a variety of datasets (\autoref{sec:global}).

% Table generated by Excel2LaTeX from sheet 'Sheet1'
% \setlength{\tabcolsep}{20pt}
\begin{table}[t]
 \small 
% \normalsize
  \centering
    \begin{tabular}{lr}
    \toprule
    \textbf{Aspects} & \multicolumn{1}{r}{\textbf{Numbers}} \\
    \midrule
    Tasks/Languages & 142/331 \\
    Features/Prompts & 318/1007  \\    
    Plain/Diagnostic Datasets  & 1,715/3,583 \\
    Annotated Datasets & 728 \\    
    Annotated Samples & 139,570,057 \\
    % Organizations & 794 \\  
    % Plain/Diagnostic Datasets  & 1,300/3,583 \\
    % Scientific datasets/papers & 415/68,189 \\
    % Annotated Datasets & 313 \\    
    % Annotated Samples & 139,501,868 \\
    % Annotated Samples & 119,890,472 139,501,868 \\
    \bottomrule
    \end{tabular}%
      \caption{Key statistics of \toolname. ``Diagnostic Dataset'' refers to a dataset obtained by applying transformations to the original version.\footnotemark  ``Annotated'' indicates datasets or samples where we compute features to obtain additional information that is not originally present in the dataset.
      %\pfliu{some numbers should be updated?} \jlfu{TBU!}
      }
  \label{tab:keystats}%
\end{table}%

With the above use cases in mind, \toolname focuses on the following design principles:

% The following text is the footnote of Tab. \ref{tab:keystats}
\footnotetext{We collect diagnostic datasets by performing an extensive literature review and searching for existing works that released diagnostic samples from different tasks.}

\begin{itemize*}
    \item \textbf{Broad-coverage}:  \toolname is designed to cover the majority of NLP tasks, and imports data from a very large number of plain datasets and diagnostic ones as shown in \autoref{tab:keystats}.\footnote{Details can be found in Appendix.} %\ref{sec:datalab-statistics}
    \item \textbf{Interpretable}: \toolname has annotated statistical information for many datasets (728 datasets,
 139,570,057 samples) 
%  \footnote{We spent months calculating these features which are all accessible in \toolname, and are still expanding. \gn{I understand that you want to emphasize that this was a big undertaking, but I'm not sure this adds a whole lot to the paper. Hopefully the message gets across without having to say this?}})
 that is not originally included in the dataset. These features can help researchers and developers better understand datasets before use, and help data creators improve data quality (e.g., removing artifacts, bias)
    \item \textbf{Unified}: One of the main goals of \toolname is to unify different data analysis and processing operations into one platform and SDK. To achieve this goal, we design a generalized typology for data and operations (\autoref{fig:typology}).
    \item \textbf{Interactive}: \toolname makes data exploration, assessment, and processing more accessible and efficient (real-time search, comparison, filtering, \textbf{generation of dataset diagnostic reports}).
    % \footnote{\url{http://datalab.nlpedia.ai/#/upload_dataset}}). 
    \toolname can also be used as an off-the-shelf \emph{annotation platform} where some missing yet important crowdsourcable information can be contributed by users.
    % \gn{The following sentence could potentially be deleted.}
    % For example, the speaker and annotator demographic information of datasets.
    \item \textbf{Inspirational}: \toolname's global view of datasets makes it possible to inspire new research directions, e.g.~by (i) finding more appropriate datasets as shown in \S\ref{sec:search} (ii) tracking the global status of dataset development and identifying future directions as illustrated in \S\ref{sec:global}.
\end{itemize*}

\section{Related Work}

\paragraph{Toolkits for NLP Pipelines}
% Recently, several general frameworks around dataset processing.

There are a wealth of toolkits that support the processing of various NLP tasks, making it easier to build a composable NLP workflow. Typical examples are 
NLTK \cite{loper2002nltk},
NLPCurate \cite{clarke-etal-2012-nlp},
Stanford CoreNLP \cite{manning2014stanford},
AllenNLP \cite{gardner-etal-2018-allennlp},
SpaCy \cite{spacy2}, GluonNLP \cite{guo2020gluoncv},
DCF \cite{dcf_2021}, \cite{huggingface_2021},
{HuggingFace} \cite{huggingface_2021}.

% Some other works either attempt to reduce the cost of manual annotation by setting up a platform: LightTag \cite{LightTag_2021}, CroAno \cite{CroAno_2021},  ET \cite{ET_2021}, 
% skweak \cite{skweak_2021},
% and ActiveAnno \cite{ActiveAnno_2021}, or focus on evaluating the robustness of the model: Checklist \cite{marco_2021}, {TextFlint} \cite{wang_textflint}, {Gym} \cite{gym_2021}, OpenAttack \cite{OpenAttack_2020}, and {TextAttack} \cite{textAttack_2020}.
In contrast to these toolkits, \toolname focuses on data analysis, bias diagnostics, and standardization of data-related operations. Moreover, besides providing the SDK, \toolname also provides a web-based interactive platform, featuring hundreds of datasets and millions of additional annotations w.r.t.~diverse features.
KYD \cite{kyd} also provides a web platform for data analysis but it mainly focuses on image data.
ExplainaBoard \cite{liu-etal-2021-explainaboard} presents an analysis platform while it focuses on system diagnostics.

\paragraph{Standardization by Community Wisdom}
In ML in general and NLP in particular, researchers have been paying increasing attention to analyzing and improving systems from the perspective of data.
In NLP, one major challenge in data processing is the diversity of data formats (e.g., \textit{CONLL}, \textit{BRAT}), task types (e.g., classification, generation) and design considerations (e.g., which types of preprocessing or augmentation) hinders the establishment of a unified platform.
Recently, however, researchers in the field are actively trying to alleviate this problem by allowing community members to collectively contribute data-related operations on the same set of code frameworks, and eventually build a data processing platform around those operations.
For example, {HuggingFace} \cite{huggingface_2021} and Tensorflow \cite{tfdataset} Datasets, where researchers in the community contribute data loaders for different tasks and datasets.
In {XL-Augmentor} \cite{dhole2021nl} and {Prompt Sourcing} \cite{sanh2021multitask} different data transformations or prompts are crowdsourced respectively.
% {Snorkel} \cite{snorkel_2020}, for example, has different people contributing a set of data labeling operations to generate supervised data samples.
% Similarly, the Checklist platform for data perturbation, {Gym} \cite{gym_2021} and {Textflint} \cite{wang_textflint} for robustness testing.

After seeing this implicit pattern, we ask, can we have a more general 
platform
%\textbf{platform of platforms}
%\gn{I'd be OK with just ``platform'', ``platform of platforms'' seems a little bit complicated.}\pfliu{ok}
above to unify all of these different operations? \toolname makes a step towards this goal by not only focusing on how to unify data loader interfaces like Huggingface and Tensorflow have done, but also unifying data operations and analysis.

\section{\toolname}
% Not only does \toolname come with a web platform but is also has a python-supported SDK, both of which give it following unique advantages. We detail functionalities of \toolname below.

In this section we detail four major varieties of functionality provided by \toolname.

\begin{table*}[!htb]
\vspace{-3mm}
\small
  \centering 
  \renewcommand\tabcolsep{3pt}
\renewcommand\arraystretch{1.6}  
    \begin{tabular}{ccccc}
    \toprule
    \textbf{Aspect} & \textbf{Functionality} & \textbf{Input} & \multicolumn{2}{c}{\textbf{Example Output}} \\
    \midrule
    % Data Diagnostics
    \multirow{9}[2]{*}{Diagnostics}          & \multirowcell{3}{Fine-grained \\ Analysis} & \multirowcell{3}{One dataset} & \multirow{3}[1]{*}{\includegraphics[scale=0.37]{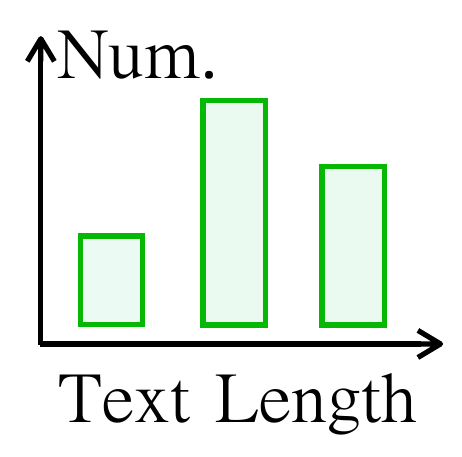}} & \multirow{3}[1]{*}{\makecell[{{p{6cm}}}]{\textbf{Characteristic histogram}: The bar chart on the left shows the distribution of the number of samples of different lengths in a dataset. 
         %\gn{Maybe this should be before ``bias analysis''? It seems more ``basic'' and also is listed first in section 3.1.}
         }} \\ \\ \\
         \cmidrule(lr){2-5}
& \multirowcell{3}{Bias Analysis} & \multirow{3}[1]{*}{One dataset} & \multirow{3}[1]{*}{\includegraphics[scale=0.4]{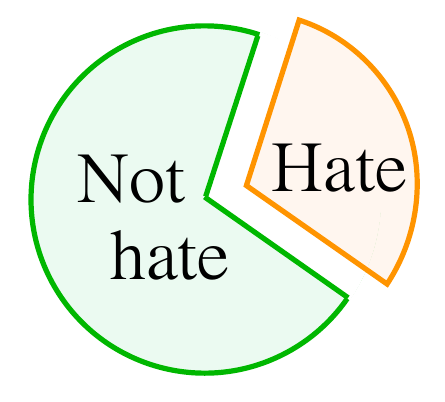}} &
    \multirow{3}[1]{*}{\makecell[{{p{6cm}}}]{\textbf{Bias pie chart}: The pie chart on the left shows the hate speech bias of a dataset. The orange portion on the right is the percentage of the data samples containing hate speech, while the green portion on the left is the rest.}}
                        \\ \\ \\         
          \cmidrule(lr){2-5}
          & \multirowcell{3}{Interactive \\ Analysis}& \multirow{3}[1]{*}{Multiple datasets} & \multirow{3}[1]{*}{\includegraphics[scale=0.37]{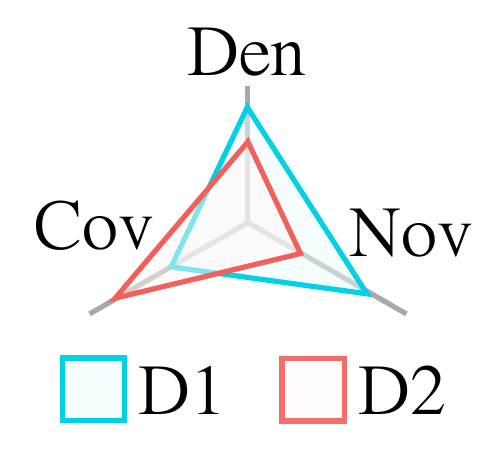}} & \multirow{3}[1]{*}{\makecell[{{p{6cm}}}]{\textbf{Comparison spider chart}: The spider diagram on the left shows the differences between two datasets (D1 and D2) in three dimensions: \texttt{Den}: density, \texttt{Nov}: novelty, \texttt{Cov}: coverage.}.}
          \\ \\ \\
          \midrule

       \multirow{3}{*}{Operations} & \multirowcell{3}{Aggregate, Edit,\\ Featurize, Prompt,\\Preprocess} & \multirowcell{3}{One dataset} & \multirow{3}{*}{\includegraphics[scale=0.28]{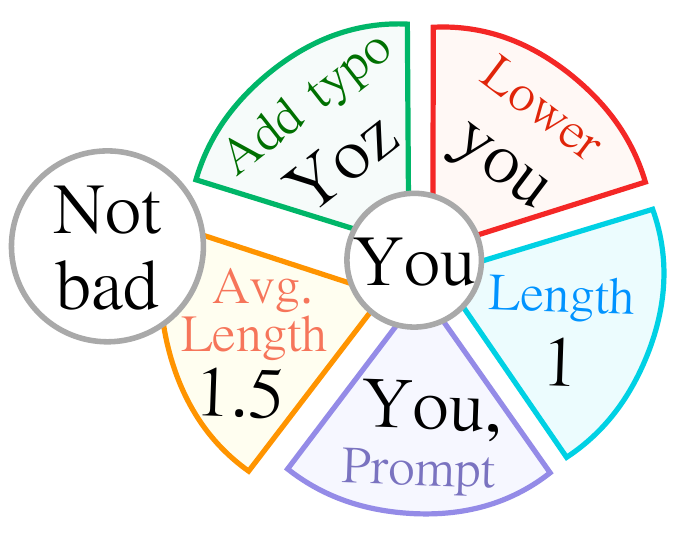}} & \multirow{3}{*}{\makecell[{{p{6cm}}}]{\textbf{Statistics or transformed datasets}: The figure here shows five example operations (one for each category) computed on either one sample (\texttt{You}) or the whole dataset (\texttt{You} and \texttt{Not bad}).}} \\ \\ \\
       \midrule
       
        % Data Search
       \multirow{3}{*}{Data Search} & \multirowcell{3}{Search} & \multirowcell{3}{Keywords, \\textual descriptions,\\similar dataset} & \multirow{3}{*}{\includegraphics[scale=0.3]{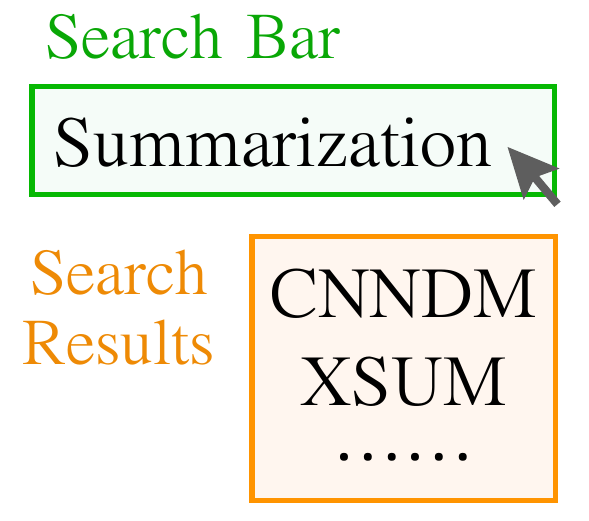}} & \multirow{3}{*}{\makecell[{{p{6cm}}}]{\textbf{Related datasets}: The example on the left uses the keyword \texttt{Summarization} to search for recommended datasets. 
       %The returned datasets are highly relevant \gn{Do we need this last sentence? It's definitely not always true.}
       .}} \\ \\ \\
       \midrule

       % Global Vision
         \multirow{3}{*}{Global Vision} 
            & \multirowcell{3}{Language Map} & \multirowcell{3}{Multiple datasets} & \multirow{3}{*}{\includegraphics[scale=0.55]{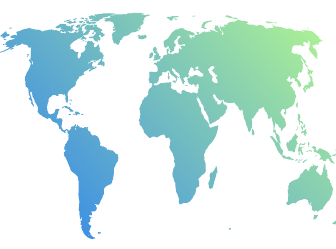}} & \multirow{3}{*}{\makecell[{{p{6cm}}}]{\textbf{Heatmap}: We use a heatmap 
            %\gn{``heat map'' or ``heatmap'', be consistent. Either is fine with me.}
            to show how many datasets are available for each country in terms of the languages people speak in that country.}} \\ \\ \\
        %   \cmidrule(lr){2-5}
        %   & \multirowcell{3}{Performance Map} & \multirow{3}[1]{*}{Multiple datasets} & \multirow{3}[1]{*}{\includegraphics[scale=0.55]{fig/global_vision_analysis/performance_map.pdf}} & \multirow{3}[1]{*}{\makecell[{{p{6cm}}}]{\textbf{Heat map of performance distribution}: The heat map on the left shows the distribution of system performance for different countries on the same task(s).}}
  
    \bottomrule
    \end{tabular}%
      \caption{A graphical breakdown of the functionality of \toolname. \
      %\jlfu{Have we forgotten about online diagnostics?} %\pfliu{we don't explicitly reflect it here.}
      }
  \label{tab:keyfunctionalities}%
\end{table*}%

\subsection{Data Diagnostics}
\label{sec:diagnostics}

Data diagnostics aim to provide users with a comprehensive picture of data through various statistical analyses, enabling better model designs.

\subsubsection{Fine-grained Analysis}

Fine-grained analysis aims to answer the question: \textit{what are the characteristics of a dataset?}
Existing works have shown its advantages in better system designs~\cite{zhong-etal-2019-closer,fu-etal-2020-rethinkcws,tejaswin-etal-2021-well} . 
Conceptually, this analysis over various dimensions can be performed over each data point (i.e.~sample-level) or whole datasets (i.e.~dataset-level).
These are either generic (\textit{text length} at sample-level or \textit{the average text length} at corpus-level) or task-specific (for summarization: \textit{summary compression} \cite{chen-etal-2020-cdevalsumm} or \textit{the average of summary compression}). We detail the features utilized for fine-grained analysis in Appendix. % \ref{app:feagures}
 
One key contribution of \toolname is that we not only design rich sample-level and dataset-level features, but also compute and store those features in a database for easy browsing.
% \footnote{We use MongoDB to store all data and features, which makes it easier for users to interact with the web platform.\gn{The connection between MongoDB and ``easy to interact'' will not be very clear if you don't know much about MongoDB. Either clarify or delete this footnote, where I slightly prefer the latter.}} 
As shown in \autoref{tab:keystats}, so far, we have designed more than 300 features and computed features for 140M samples.

\subsubsection{Bias Analysis} 
The research question to be answered by bias analysis is: \textit{Does the dataset contain potential bias (e.g., artifacts, gender bias)?}
Bias problems have been discussed extensively in NLP \cite{zhao_2018,blodgett-etal-2020-language}, and we argue that establishing a unified platform for data bias analysis can more efficiently identify or prevent (for data creators) data bias problems. For example, through the artifact analysis, users can know the shortcut provided by the dataset for model training and be inspired to design more robust systems.
%\gn{Based on the taxonomy of \citet{blodgett-etal-2020-language}, what kinds of harms due to bias would DataLab be able to solve?}.
So far, \textsc{DataLab} supports three types of bias analysis.

%\paragraph{Pointwise Mutual Information (PMI)}
\paragraph{Artifact Identification}
As observed in many previous works \cite{Gururangan_2018,mccoy-etal-2019-right},
artifacts commonly exist in datasets, which provide shortcuts for model learning and therefore reduce its robustness.
\toolname allows researchers to easily identify potential artifacts in a dataset using the features we have pre-computed for each sample.
Specifically, we use PMI (Pointwise mutual information) \cite{bouma2009normalized} to detect whether there is an association between two features (e.g. sentence length vs. label). We detail this method using an example in Appendix.\footnote{\url{https://expressai.github.io/DataLab/docs/WebUI/bias_analysis_for_artifacts}} % \ref{app:bias}

\begin{figure*}
    \centering
    \subfloat[\centering Data]{{\includegraphics[width=0.35\linewidth]{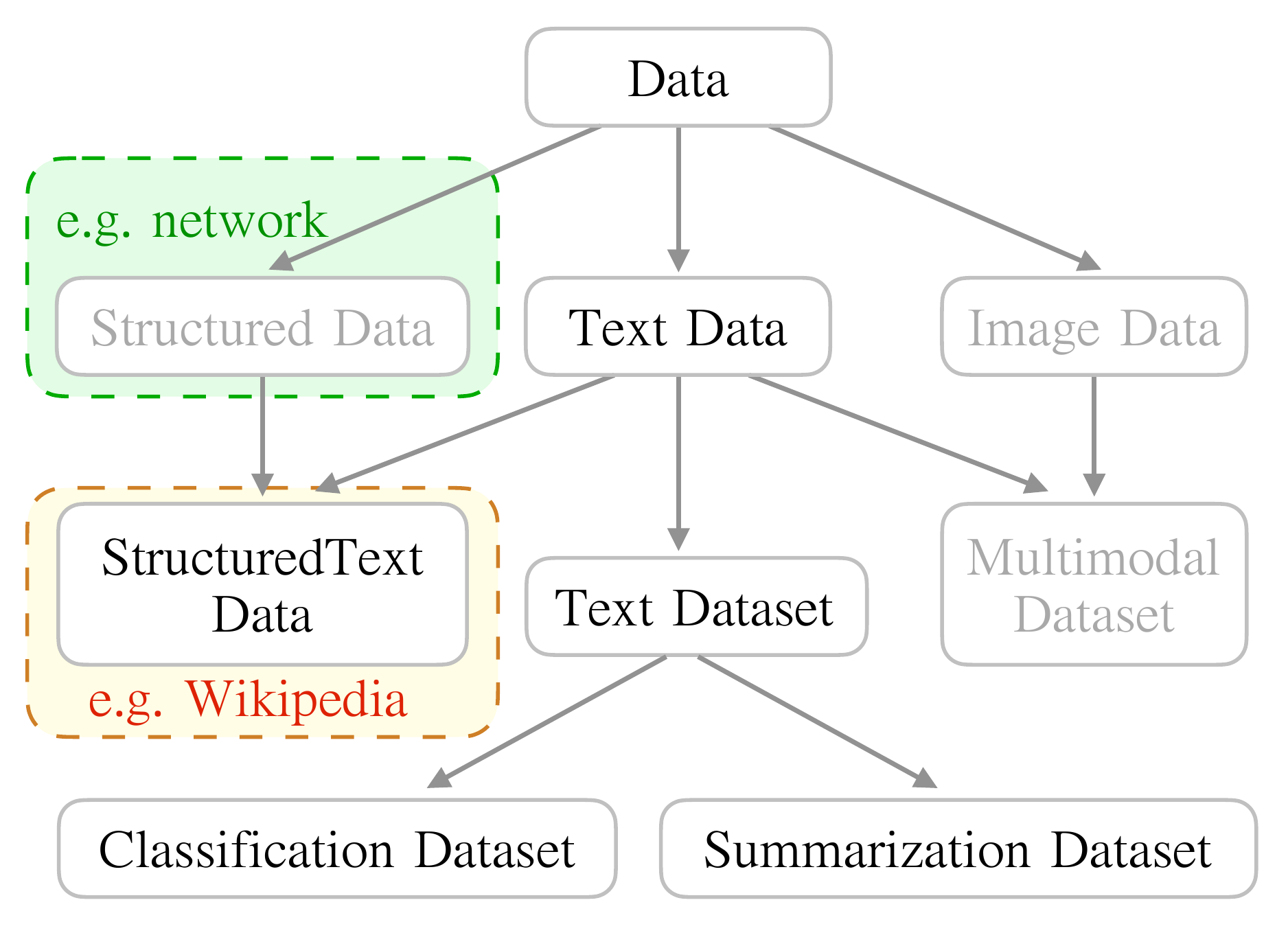} }}%
    \hspace{1mm}
    \subfloat[\centering Operation]{{\includegraphics[width=0.29\linewidth]{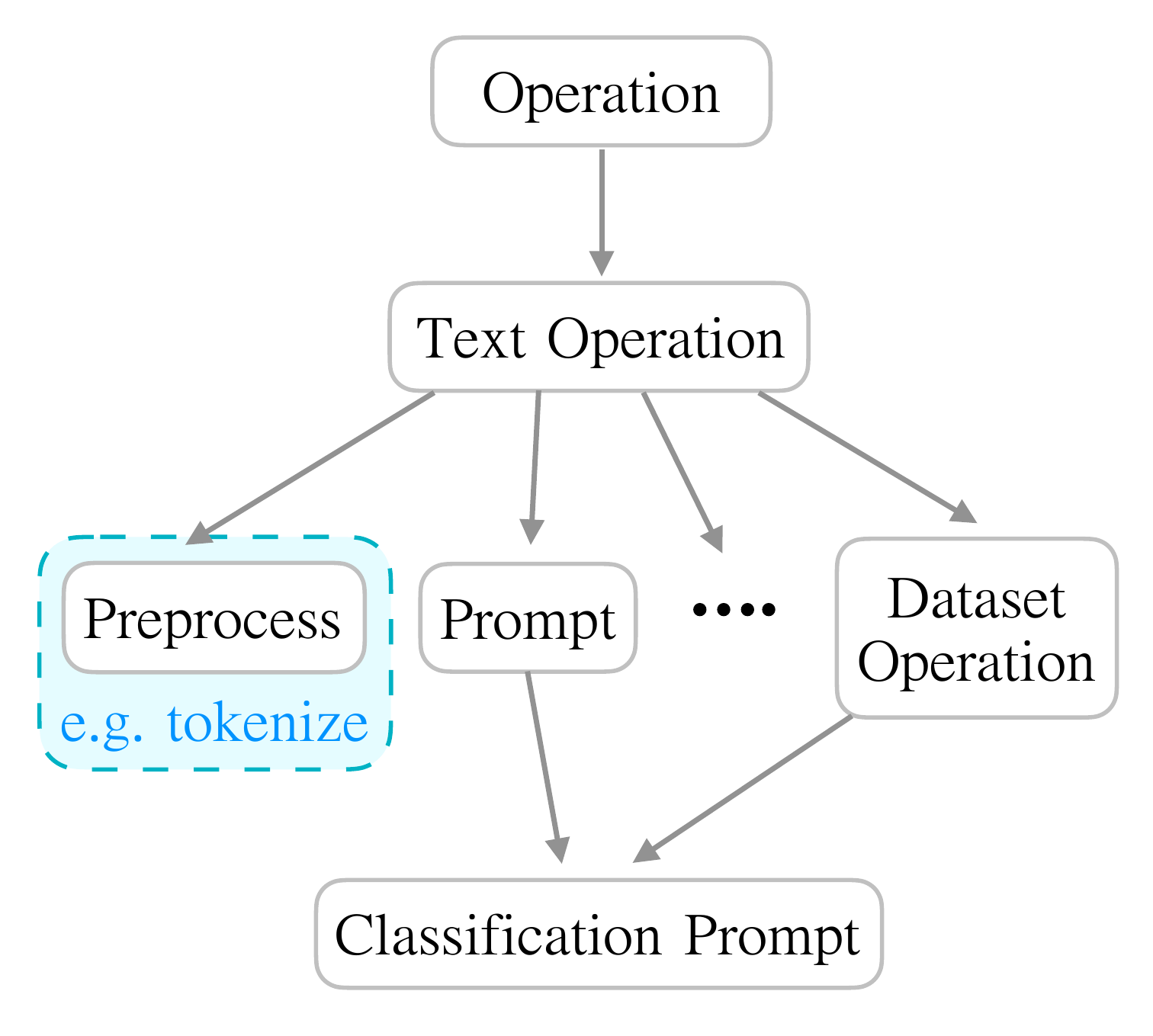} }}%
    \hspace{1mm}
    \subfloat[\centering SDK]{{\includegraphics[width=0.3\linewidth]{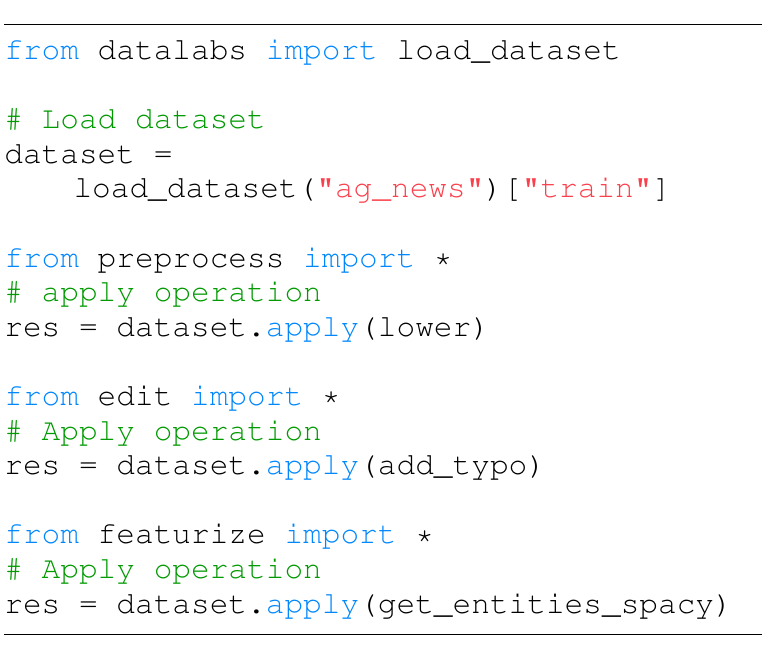} }}%
    \caption{Typology of Data and Operations. Gray-white text (e.g., \textit{Image Data}) indicates that the data type has been defined but we have not yet added data of that type.}    
    \label{fig:typology}
\end{figure*}

\paragraph{Gender Bias Analysis} 
Gender bias is a prevalent social phenomenon. 
% Training a model using a dataset containing gender bias is likely to affect the performance of that model in certain demographic group.
In this work, we introduce a multidimensional gender biased dictionary\footnote{\href{https://huggingface.co/datasets/md_gender_bias}{huggingface.gender\_bias}} used by \citet{Dinan_2005} to measure the degree of gender bias in a dataset. 
Given a dictionary $D_1$ of female names and a dictionary $D_2$ of male names.
Suppose a dataset A with $N$ samples has $n_1$ name appearing in $D_1$ and $n_2$ in $D_2$. Following \citet{gender2018jieyu}, we can calculate the female bias for dataset A as $n_1/N$; the male bias as $n_2/N$.

\paragraph{Hate Speech Analysis}

Hate speech \cite{pinkesh_2017} can lead to a "dehumanizing effect" that harms people's mental health by undermining empathy \cite{tsesis2002destructive}.
We make a first step by following \citet{thomas_2017}, classifying the samples into \textit{hate speech}, \textit{offensive language} and \textit{neither} categorizing by the ``hatesonar'' tool.\footnote{\href{https://pypi.org/project/hatesonar/}{pypi.org.hatesonar}}
We also averaged the offensiveness of all samples in a dataset to analyze the hate speech bias of the dataset.\footnote{Note that deciding whether a sentence contains toxic language is a complex task, which may involve the confounding effects of dialect and the social identity of a speaker \citet{sap-etal-2019-risk}, and future iterations of DataLab may use meta-data of datasets to further perform this analysis intersectionally. We have also stored the results of hate speech detector for all samples to make the analysis process more transparent and well-grounded and users could browse them and report error cases.}

\subsubsection{Interactive Analysis}
Interactive analysis aims to meet users' customized data analysis requirements in real time.
Although interactivity is present in many aspects of \toolname, we highlight here its use in three scenarios that make data analysis more accessible.
(1) Users can choose two datasets they are interested in and align them for comparative analysis over different dimensions, as shown in \autoref{tab:keyfunctionalities}.
(2) Users can upload their own datasets and \toolname will generate diagnostic reports for comprehensive analysis and evaluation of the datasets.
(3) Users can contribute some missing metadata information %\gn{``metadata''?}
by directly editing in the web interface.
% For example, the annotator demographic information of a dataset \gn{This is just one example of many many possibilities right? Just having one example here perhaps makes it feel a bit unnecessarily narrow.}. 

% \subsection{Diagnostic Test}

\subsection{Data Operations}
\label{sec:operations}

% What are the common ``X'' operations for text data in general and task-specific datasets?

Another key feature of \toolname is the standardization of different data operations into a unified format to satisfy different data processing requirements in one place.
To this end, we devised a general typology for the concepts of data and operation as shown in \autoref{fig:typology} and curated schemas for these objects.
% For example, when designing the schema for \textit{Text Dataset}, we combined these schemas:
% %\url{https://docs.google.com/spreadsheets/d/1bdjgv-5PHe-jjxqqHRYuBCpnDqwPFaWD5HCqL7Yep_g/edit#gid=285687177}
% (1) LREC\footnote{\href{https://lrec2020.lrec-conf.org/en/shared-lrs/}{lrec2020.en.shared-lrs}}
% (2) Data Statements \cite{bender-friedman-2018-data}, (3) Huggingface Datasets \cite{huggingface_2021} (4) Paperswithcode\footnote{\href{https://paperswithcode.com/datasets}{paperswithcode.datasets}} and merged the repetitive fields.
For the operation schema, we introduced 
(i) ``operation id'': so that researchers can report them papers for easy re-implementation for follow-up research.
(ii) ``contributor`` to credit those who contributed to the operation.
Notably, user-defined operations are also supported (we give an example in Appendix).  % \ref{app:customized_operations}

% \subsubsection{Preprocessing: What are the common preprocessing operations for text data in general and task-specific data sets?}
%\subsubsection{Preprocessing}
\paragraph{Preprocessing}
Data preprocessing (e.g., tokenization) is an indispensable step in training deep learning and machine learning models, and the quality of the dataset directly affects the learning of models.
Currently, \toolname supports both general preprocessing functions and task-specific ones, which are built based on different sources, such as SpaCy \cite{spacy2}, NLTK \cite{loper2002nltk}, Huggingface tokenizer.\footnote{\href{https://github.com/huggingface/tokenizers}{huggingface.tokenizers}}

% \paragraph{Lower Case}
% \paragraph{Tokenize}
% \paragraph{Stem}
\paragraph{Editing}
% \subsubsection{Editing: What are the popular editing operations for text data and task-specific datasets? }

Editing aims to apply certain transformations to a given text, which spans multiple important applications in NLP, for example (i) adversarial evaluation \cite{marco_2021, wang_textflint}, which usually requires diverse perturbations on test samples to test the robustness of a system. (ii) Data augmentation \cite{jason_2019,dhole2021nl,feng_2021}. Essentially, many of the methods for constructing augmented or diagnostic datasets involve some editing operation on the original dataset (e.g., named entity replacement in diagnostic dataset construction \cite{marco_2021}, token deletion in data augmentation \cite{jason_2019}).
\toolname provides a unified interface for data editing and users can easily apply to edit the data they are interested in.

\paragraph{Featurizing}
% \subsubsection{Featurizing: What are the common feature extractors for text data in general and task-specific datasets? }
This operation aims to compute sample-level features of a given text.
In \toolname, in addition to designing some general feature functions (e.g. \textit{get\_length} operation calculates the length of the text.), we also customize some feature functions for specific tasks (e.g. \textit{get\_oracle} operation for the summarization task that calculates the oracle summary of the source text.).

%Our extraction of named entities utilizes the NLTK toolkit, and automatic part-of-speech tagging draws on both the NLTK and Spacy toolkits.

\paragraph{Aggregating}
% \subsubsection{Aggregating: What are the meaningful dataset-level statistics for text data and task-specific datasets?  }
Aggregation operations are used to compute corpus-level statistics such as TF-IDF \cite{salton1988term}, label distribution. Currently, \toolname supports both generic aggregation operations applicable to any task and some customized ones for four NLP tasks (classification, summarization, extractive question answering and natural language inference).

% To observe the features of datasets and compare the differences between different datasets, we used the \texttt{aggregating operation} to aggregate sample-level features into dataset-level features. 
% The \texttt{aggregating operation} can be, calculate the average sentence length, label distribution of a data set.

\begin{figure}[h]
    \centering
    \includegraphics[width=0.70\linewidth]{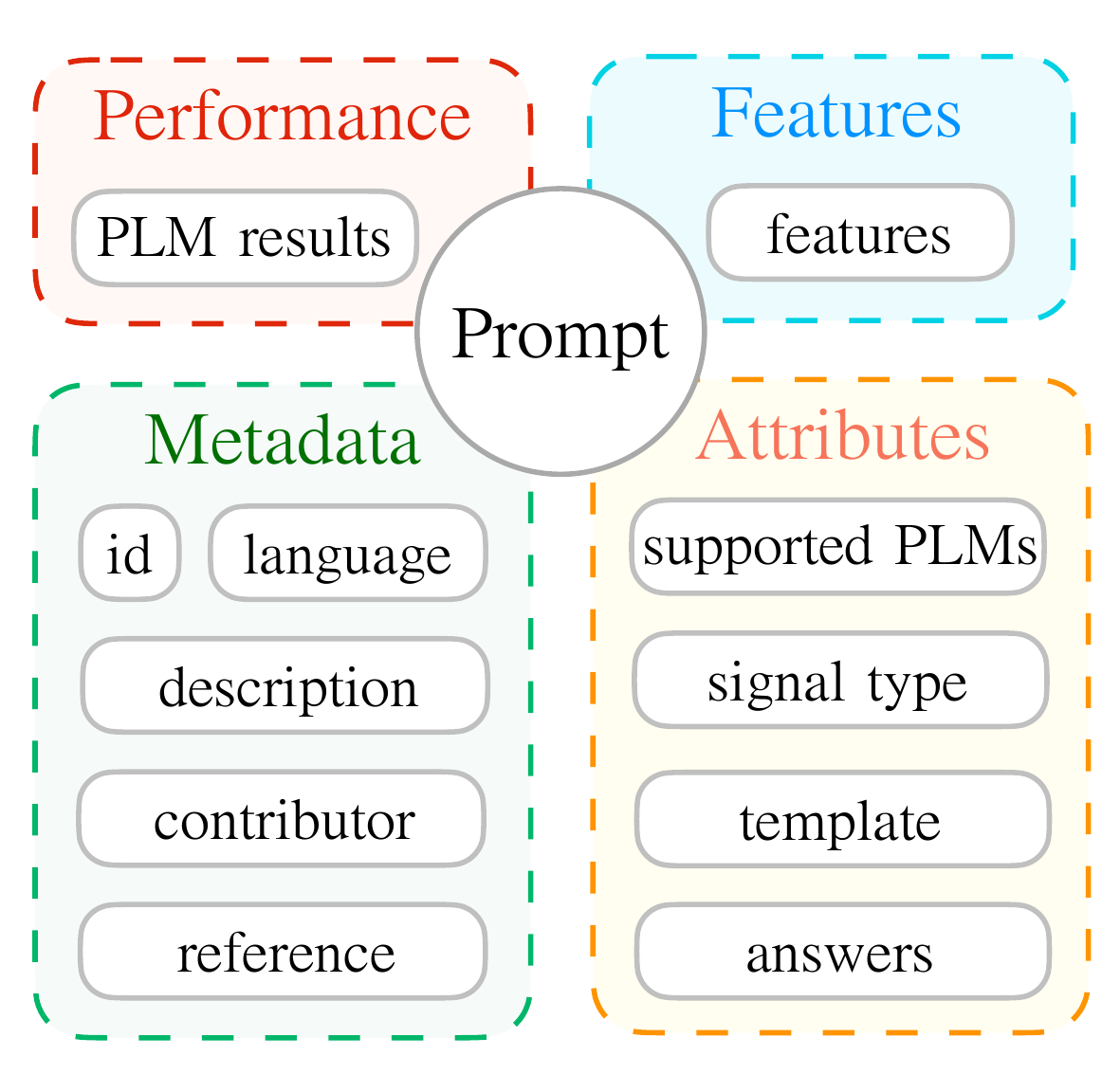}
    \caption{Prompt schema in \toolname, where ``PLM" represents the pre-trained language model.}
    \label{fig:prompt_schema}
\end{figure}

\paragraph{Prompting}
% \subsubsection{Prompting: What are suitable prompt designs for a specific dataset?}
Prompt-based learning \cite{liu_2021} has received considerable attention, as better utilization of pretrained language models benefits many NLP tasks.
In practice, what makes a good prompt is a challenging question. We define the prompt schema as shown in fig.~\ref{fig:prompt_schema}. The elements we included in a prompt cover diverse aspects including its features (e.g. length, shape, etc.), metadata (e.g. unique identifier, language, etc.), attributes (e.g. template, answers, etc.) as well as its performance w.r.t. different pre-trained language models and settings. The design can not only help researcher design prompts but also analyze what makes a good prompt.

So far, \toolname covers 1007 prompts which can be applied to five types of tasks (topic classification, sentiment classification, sentence entailment, summarization, natural language inference), covering 309 datasets in total. 

\subsection{Data Search}
\label{sec:search}
Data search aims to answer the research question: which datasets should one use given a description of a research idea? 
As more datasets are proposed, there is an open question of how to choose the right dataset for a given application.
\toolname takes a step towards solving this problem by including semantic dataset search.

\begin{figure*}
    \centering
     \subfloat[\centering PMI analysis between two features: hypothesis length and label for the SNLI]{{\includegraphics[width=0.9\linewidth]{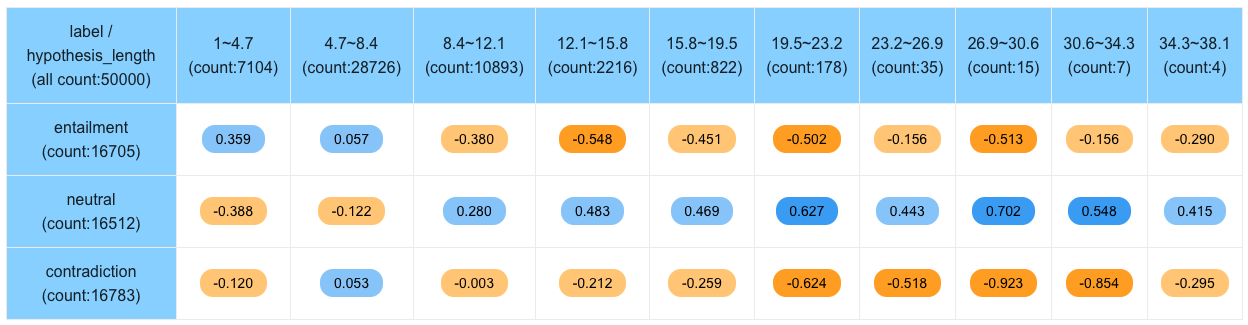} }}%
    \hspace{1mm} \\
    \subfloat[\centering Language map]{{\includegraphics[width=0.5\linewidth]{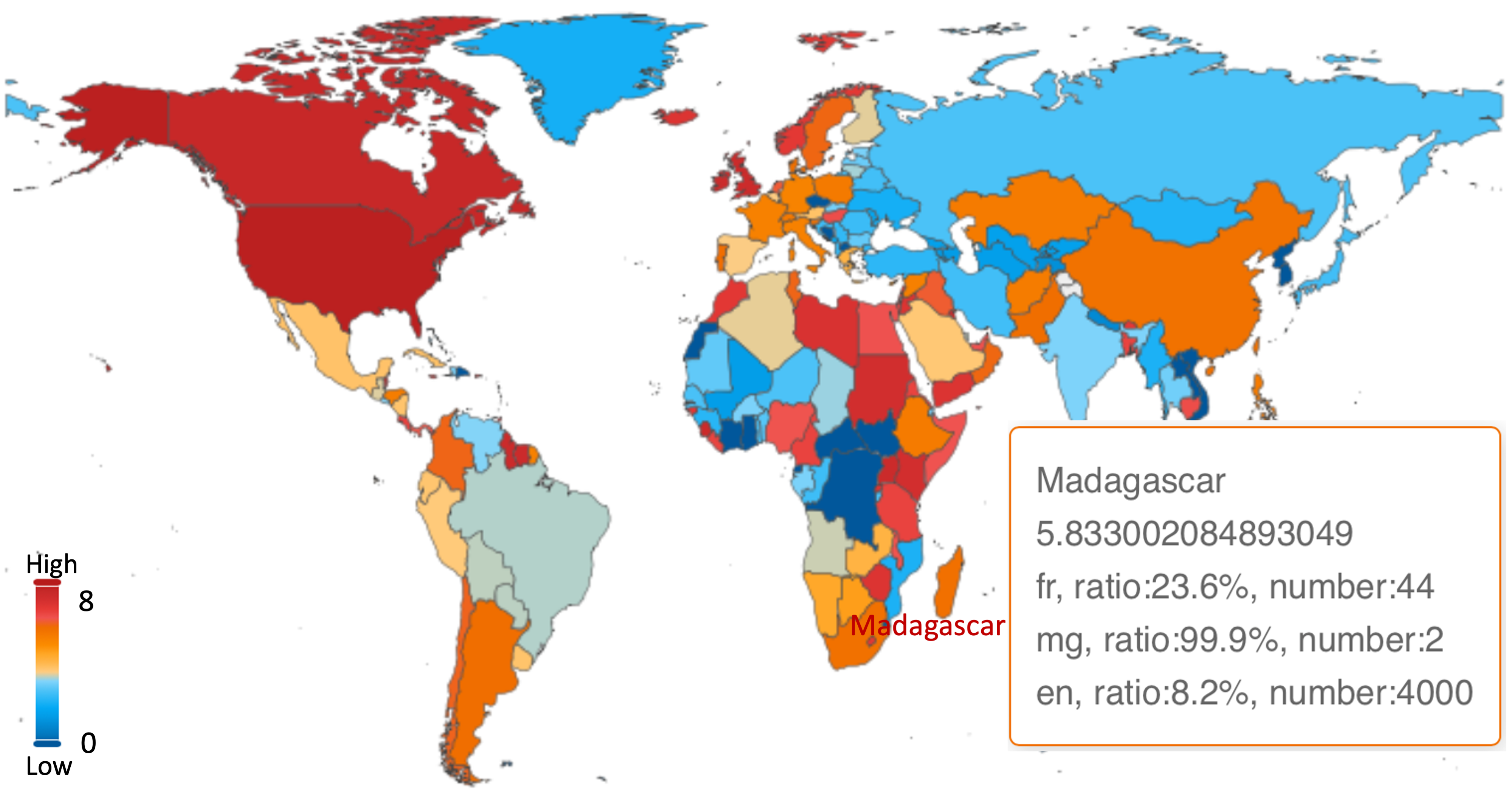} }}%
    \hspace{2mm}
    \subfloat[\centering Data recommendation]{{\includegraphics[width=0.4\linewidth]{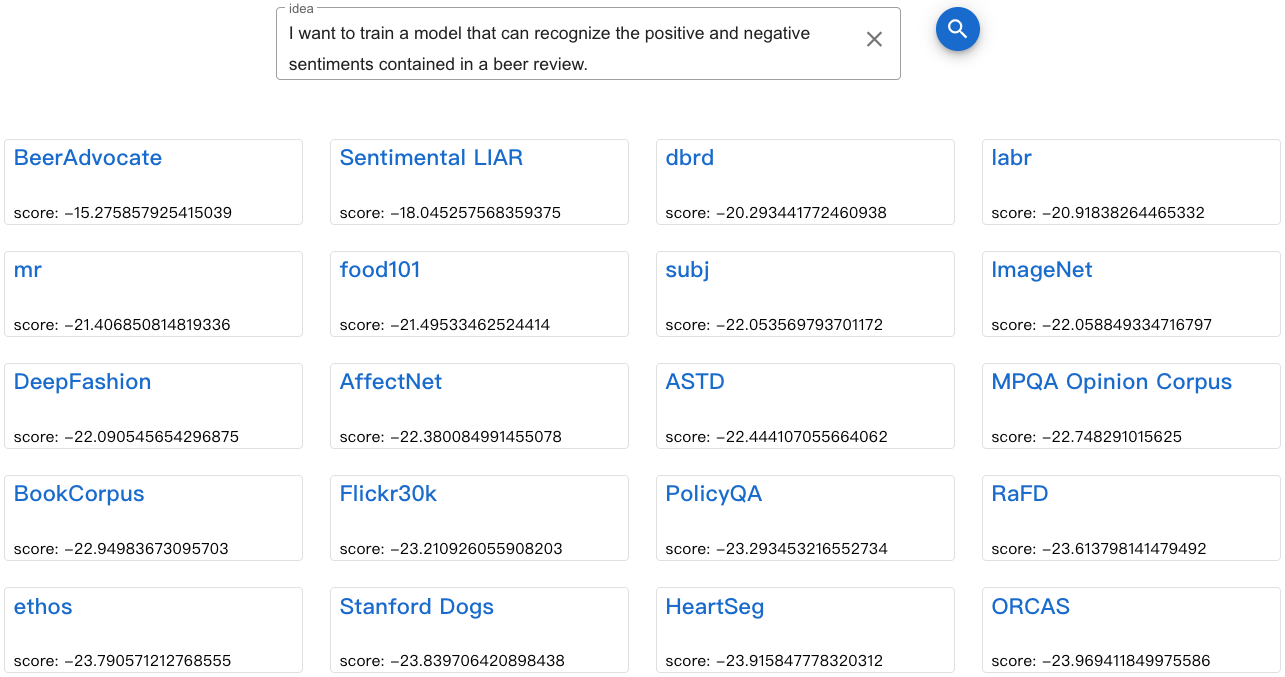} }}%
    \caption{Case studies on artifact detection, systemic inequality, and dataset recommendation. 
    %\pfliu{the figure (a) could be updated using our new UI.} \jlfu{done!}
    }    
    \label{fig:case-study}
\end{figure*}

% Fig.\ref{xx} gives an example to illustrate this.

% \subsubsection{Description-based: Which datasets should I use given a textual description of a research idea?}
\toolname data search takes a natural language description of a research idea,\footnote{\toolname also supports keyword queries as input. However we find the added context provided by natural language descriptions improves search quality.} compares it with descriptions of thousands of datasets, and displays the datasets best matching the input (a detailed example is given in  \autoref{fig:case-study}). This retrieval system goes beyond keyword search by using semantic matching. The algorithm is described in a pending paper; we provide technical details in  Appendix.
%\autoref{app:dataset_finder}

\subsection{Global Vision Analysis}
\label{sec:global}
\paragraph{Language Map: Which languages' datasets get less attention?}
A language map is used to analyze which languages are more studied and which are less studied from a geographical view \cite{faisal2021dataset}, identifying potential systemic inequalities.
Specifically, we first count how many datasets are available for each language. Then for each country we calculate a distribution over languages,\footnote{We refer to some official statistics from \href{https://www.cia.gov/the-world-factbook/countries/}{this link}.} where the ratio of each language represents the proportion of people who speak that language. Finally, for each country, we can get the weighted average number of datasets available for it in terms of its spoken languages (see Appendix for details).
%\autoref{app:lmap}

% Therefore, for each country, we can calculate a number that describes how well the country is covered by the available datasets in its spoken language.

\section{Case Study}
We perform three case studies to show the utility of \toolname and put more in the appendix.
% \pfliu{DataLab has been used for ExplainaBoard, we can also cover this?}

\paragraph{Artifacts}

% We take a natural language inference dataset \texttt{SNLI} \cite{bowman-etal-2015-large} as an example and use \toolname to identify its potential artifacts. 

One famous example of a dataset artifact reported by \citet{Gururangan_2018} (Figure 1) is that in NLI datasets, the length of the hypothesis sentence is closely associated with the assigned label of the premise-hypothesis pair.
In fact, \toolname is able to easily re-discover this artifact, and more.
Fig.~\ref{fig:case-study}-(a) shows an analysis on the SNLI dataset \cite{bowman-etal-2015-large} between two features {length}$_{\text{hypothesis}}$  and {label} (entailment, neutral or contradiction). 
We can observe that, when {length}$_{\text{hypothesis}}$ is larger than 8.4, $\mathrm{PMI}(\text{label}_{\text{neutral}}, \text{length}_{\text{hypothesis}} )>0.28$, suggesting that ``long hypotheses'' tend to co-occur with the ``neutral'' label, even without consideration of the premise.
Additionally, when {length}$_{\text{hypothesis}} \in [1,4.7]$, $\mathrm{PMI}(\text{label}_{\text{entailment}}, \text{length}_{\text{hypothesis}})=0.359$, implying that
``short hypotheses'' tend to co-occur with the label ``entailment''.
However, this is not all; \textbf{we further observed more than ten potential artifacts on \texttt{SNLI} and another popular dataset \texttt{SST2}} \cite{socher-etal-2013-recursive} (see Appendix), which demonstrates the ability of \toolname to efficiently identify these artifacts.

\paragraph{Systemic Inequalities}
% We show the extent to which language has been studied from a global perspective. 
Fig.~\ref{fig:case-study}-(b) is a statistic of the degree to which languages are studied from a global (w.r.t each country in the world) perspective, with a darker red indicating more datasets studied/constructed for the languages spoken in a given country, and darker blue indicating the opposite.
Unsurprisingly, we observe that \textit{English} is the most studied (large English-speaking countries like the US, Canada, and UK are in dark red), which also benefits those English-speaking African countries (e.g. 
Madagascar, 
Uganda,  
and Libya are in red.). We also observe that the languages spoken in
\texttt{bm} (Mali), \texttt{ee} (Ghana), and \texttt{kr} (Niger) are rarely studied, as can be seen from our language map that these three languages have a value of 0.

\label{app:additional_case_study}
\begin{figure}[!t]
    \centering
    \includegraphics[width=0.6\linewidth]{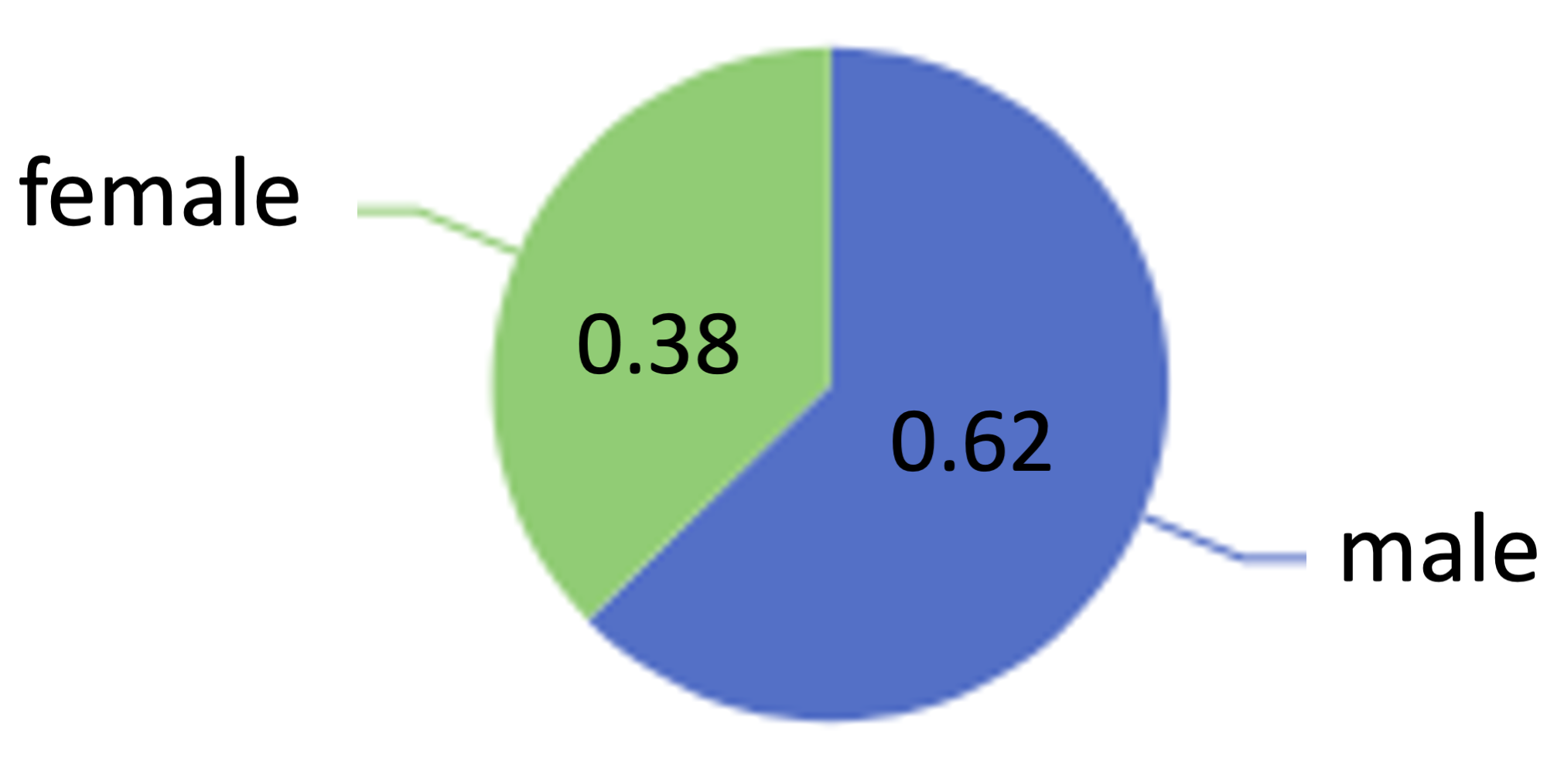}
    \caption{Gender bias analysis on SNLI.}
    \label{fig:gender_bias}
\end{figure}

\paragraph{Gender Bias}
We also showcase the gender bias analysis on \texttt{SNLI} as illustrate in Fig.~\ref{fig:gender_bias}.
%Our bias analysis is based on the proportion of male words (female words) contained in the sample to the total number of samples.
We can find that the samples in the  \texttt{SNLI} dataset contain more male-oriented words than females ($\text{male}(0.62)>\text{female}(0.38)$).

\paragraph{Dataset Recommendation}
% \pfliu{we can change the example: \url{https://github.com/ExpressAI/DataLab_dev/issues/101}} \jlfu{done!}
% To make academic research easier, our system provides a service to recommend suitable datasets for users' ideas. 
Fig.~\ref{fig:case-study}-(c) presents a case study of using our \toolname to get recommended datasets.
When a user enters a research idea ``\textit{I want to train a model that can recognize the positive and negative sentiments contained in a beer review.}'', \toolname returns the beer review dataset \texttt{BeerAdvocate} \cite{mcauley2013amateurs} first in the interface, which is a precise result since the dataset consists of beer reviews from beeradvocate.\footnote{https://www.beeradvocate.com/}
% can reasonably recommend sentiment analysis datasets \gn{I don't think the follwoing datasets are correct: the query above returns BeerAdvisor in the interface.}, such as \texttt{BeerAdvocate} \cite{mcauley2013amateurs}
% such as \texttt{BeerAdvocate} (Arabic Sentiment Tweets Dataset) \citep{DBLP:conf/emnlp/NabilAA15}, \texttt{MPQA} \citep{DBLP:journals/lre/WiebeWC05}, and \texttt{Sentimental LIAR} \citep{DBLP:conf/isi/UpadhayayB20}.

\section{Implications and Roadmap}

\toolname was born from our two visions
(1) It is essential to standardize both the format of data and the interface of data-centric operations.
% For example, it helps us to establish a unified platform for data analysis and processing.
(2) The standardization of data and operations allows more people in the community to contribute and share community wisdom. For example, in \toolname, community researchers can easily contribute (1) new  feature functions that enable us to conduct data analysis from more dimensions; (2) new datasets or the missing metadata.
We hope that the unity of the platform can make it easier for \textit{collective wisdom} to come into play. In the future, we will continue to expand \toolname in multiple directions: more data types (e.g., image), more operations (e.g., labeling function~\cite{snorkel_2020}), promoting better progress in the field.

\section{Ethics/Broader Impact Statement}

If the platform works as expected, researchers, developers and analysts can all benefit from it. Researchers can gain a deeper and broader understanding of the characteristics of the datasets, developers can more easily access the datasets and manipulate the data samples, and analysts can see some social insights from the datasets.

During the whole data analysis process, we tried to make it as transparent as possible and the results of the analysis were well-grounded on the sufficient evidence so that users can more reliably use it. Additionally, uses are encouraged to report the case where the annotation results are not precise.
Currently, \textsc{DataLab} only supports public datasets. In addition, knowing more about the characteristics of the test sets might make overfitting easier for model training. One possible approach is through multi-dataset evaluation, i.e., a good system should achieve good results across a series of different datasets.

\section*{Acknowledgements}
We thank Antonis Anastasopoulos for sharing the mapping data between countries and languages, and thank
Alissa Ostapenko, Yulia Tsvetkov, Jie Fu, Ziyun Xu, Hiroaki Hayashi, and Zhengfu He for useful discussion and suggestions.
This work was supported in part by Grant No.~2040926 of the US National Science Foundation, and the National Science Foundation of Singapore under its Industry Alignment Fund – Pre-positioning (IAF-PP) Funding Initiative. Any opinions, findings, conclusions, or recommendations expressed in this material are those of the authors and do not reflect the views of the National Research Foundation of Singapore and the US National Science Foundation.

% Entries for the entire Anthology, followed by custom entries
\bibliography{anthology,custom}
\bibliographystyle{acl_natbib}

\appendix

\section{Appendix}
\label{sec:appendix}

\subsection{Detailed Statistics of \toolname.}
\label{sec:datalab-statistics}
Here, we list more detailed statistics of \toolname in \autoref{tab:app-detail-stats}.

\begin{table}[!htb]
  \centering \small
  \renewcommand\tabcolsep{3.5pt}
    \renewcommand\arraystretch{1.3}  
    \begin{tabular}{llc}
    \toprule
    \multicolumn{2}{l}{\textbf{Aspect}} & \multicolumn{1}{l}{\textbf{Number}} \\
    \midrule
    \multicolumn{2}{l}{Tasks} & 142 \\
    \multicolumn{2}{l}{Plain datasets} & 1,715 \\
    \multicolumn{2}{l}{Diagnostics datasets} & 3,583 \\
    \multicolumn{2}{l}{Language} & 331 \\
    \multicolumn{2}{l}{Organization} & 794 \\
    \multicolumn{2}{l}{Prompts} & 1,007 \\
    \cdashline{1-3}[0.8pt/2pt]
    \multirow{5}[0]{*}{Operation} & Aggregate & 8 \\
          & Preprocess & 4 \\
          & Featurize & 16 \\
          & Edit  & 23 \\
          & Prompt & 32 \\
    \cdashline{1-3}[0.8pt/2pt]
    \multirow{2}[1]{*}{Feature} & Sample level & 138 \\
          & Dataset level & 180 \\
     \cdashline{1-3}[0.8pt/2pt]
     \multirow{4}[1]{*}{Bias analysis}
     & Hate speech datasets & 240 \\
      & Gender bias datasets & 241 \\
     & Gender bias samples & 18,520,130 \\
      & Hate speech samples & 18,511,763 \\
     \cdashline{1-3}[0.8pt/2pt]
     
     \multicolumn{2}{l}{Annotated Datasets} & 728 \\
     \multicolumn{2}{l}{Annotated samples} & 139,570,057 \\
     \multicolumn{2}{l}{Total samples} & 408,460,905 \\
    %  \multicolumn{2}{l}{Total samples} & 408,392,716 \jlfu{add the number of scientific papers} \\
     
    \bottomrule
    \end{tabular}%
    \caption{More detailed statistics of the \toolname. ``Diagnostic Dataset'' refers to a dataset obtained by applying transformations to the original version.\footnotemark  ``Annotated'' indicates datasets or samples where we compute features to obtain additional information that is not originally present in the dataset.}
  \label{tab:app-detail-stats}%
\end{table}%

%  \caption{Key statistics of \toolname. ``Diagnostic Dataset'' refers to a dataset obtained by applying transformations to the original version.\footnotemark  ``Annotated'' indicates datasets or samples where we compute features to obtain additional information that is not originally present in the dataset.}

\subsection{Features}
\label{app:feagures}
Features (e.g., \texttt{sentence length}) allow us to understand the characteristics of a dataset from different perspectives.
Following \citet{Fu_2020}, we define 318 features for 142  NLP tasks. 
Below, we list some core features at the sample- and dataset-level and suitable tasks.

\subsubsection{Sample-level}
\paragraph{General Features}

General features are task-agnostic and suitable for all NLP tasks.

\begin{itemize*}
    \item \textbf{Sentence length}: the number of tokens in a sentence.
    \item \textbf{Part-of-speech tags}: the part-of-speech tag for each token is automatically labeled by NLTK \cite{loper2002nltk} Python tool.
    \item \textbf{Named entities}: entity names are automatically recognized by NLTK and SpaCy \citep{spacy2} Python tools.
    \item \textbf{Basic words ratio}: the proportion of words that appear in the basic English dictionary\footnote{\href{https://simple.wikipedia.org/wiki/Wikipedia:List_of_1000_basic_words}{wikipedia.basic\_words}}.
    % \item \textbf{Gender bias}: the proportion of words containing male/female in sentences was computed based on X1, X2, and X3 gender-biased dictionary.
    \item \textbf{Lexical richness} \citep{richards_1987}: the proportion of unique words, obtained by dividing the number of unique words by the total number of words.
    \item \textbf{OOV density}: the proportion of words in a test sentence that do not appear in the training set.
\end{itemize*}

\paragraph{Specialized Features}
In addition to general features, we also design task-specific features for some core NLP tasks.
Below, we list some key task-specific features, as well as applicable tasks.

\begin{itemize*}
% for text classification
% for ner task
\item \textbf{Span length}: the length of span. Span can be entity/answer/chunk/aspect. (\textit{NER}, \textit{QA}, \textit{Chunking}, \textit{ABSA})
\item \textbf{Label consistency of span} \citep{Fu_2020}: the visibility of a span and its label in the training set. (\textit{NER}, \textit{Chunking})
\item \textbf{Span frequency}: the frequency of entities in the training set. (\textit{NER}, \textit{Chunking})
\item \textbf{Span density}: the number of words belonging to entities in a sentence divided by the length of the sentence. (\textit{NER}, \textit{Chunking})

% sentence pair classification..
\item \textbf{Text similarity}: measures how similar two texts are. Here, we explore BLEU \citep{Papineni_2002} and ROUGE2 \citep{lin-2004-rouge} for two texts. (\textit{SUMM}, \textit{Match}, \textit{QA})
\item \textbf{Text length comparison}: measures the sentence-length relationship of sentence pairs, including addition, subtraction, and division operation of sentence lengths. (\textit{Match}, \textit{SUMM},\textit{QA})
% for QA
\item \textbf{Answer/span position}: measures where the answer/span starts in the text. (\textit{QA}, \textit{ABSA}, \textit{Chunking})

% for summarization
\item \textbf{Coverage ratio}: measures to what extent a summary covers the content in the source text. (\textit{SUMM})
% \item \textbf{Compression ratio}: measures the compression ratio from the source text to the generated summary (\textit{SUMM})
% \item \textbf{Repetition ratio}: measures the rate of repeated segments in summaries. The segments are instantiated as trigrams. (\textit{SUMM})
% \item \textbf{Novelty ratio}: the proportion of segments in the summaries that have not appeared in source documents. The segments are instantiated as bigrams. (\textit{SUMM})
\item \textbf{Copy length}: the average length of segments in a summary copied from the source document. (\textit{SUMM})
\end{itemize*}

The full names of the tasks mentioned above are as follows:
\vspace{-5pt}
\begin{itemize*}
    % \item \textbf{TC}: Text Classification
     \item NER: Named Entity Recognition
     \item Chunking: Chunkinig
     \item POS: Part-of-speech Tagging
     \item ABSA: Aspect-Based Sentiment Analysis
     \item QA: Question Answering
     \item {Matching}: Text Matching
     \item {SUMM}: Text Summarization
\end{itemize*}

% \begin{figure}
%     \centering
%     {{\includegraphics[width=0.9\linewidth]{fig/data_typology.png} }}%
%     \caption{Typology of Data.}
%     \label{fig:my_label}
% \end{figure}

\subsubsection{Dataset-level}
% \paragraph{General Features}
\begin{itemize*}
    \item \textbf{Average on dataset-level}: a sample-level feature can be converted into a dataset-level feature by averaging that feature of each sample in the dataset (e.g. the average text length, the average span length). 
    \item \textbf{Distribution of vocabulary}: measured by the word frequency of each word in the dataset.
    \item \textbf{Distribution of label}: characterize the number of samples contained in each category in the dataset.
    \item \textbf{Sample size of different splits}: characterize the number of samples contained in different splits.
    % \item \textbf{hate speech ratio}:
    \item \textbf{Hate speech ratio}: characterize the degree of hate speech bias of the dataset. 
    \item \textbf{Spelling errors ratio}: measures the extent of spelling errors contained in a dataset with the help of a detection tool\footnote{\href{https://github.com/jxmorris12/language_tool_python}{spelling\_error\_detect\_tool}}.
    
\end{itemize*}
% \paragraph{Specialized Features}

\begin{table*}[ht]
  \centering \footnotesize
    \renewcommand\tabcolsep{2.7pt}
    \renewcommand\arraystretch{1.3}  
    \begin{tabular}{lp{8cm}}
    \toprule
    \textbf{Observation} & \textbf{Conclustion} \\
    \midrule
    \textbf{SNLI} &  \\
    \midrule
    $\text{len}_\text{hp} >8.4$, $\mathrm{PMI}(\text{label}_{\text{neutral}}, \text{len}_{\text{hp}} )>0.28$; & Long hypotheses tend to be neutral. \\
    $\text{len}_\text{hp} \in [1,4.7]$, $\mathrm{PMI}(\text{label}_{\text{entailment}}, \text{len}_{\text{hp}} )=0.359$; & Short hypotheses tend to be entailment. \\
    \cmidrule(lr){1-2}
    
    $\text{flesch\_reading\_ease}_\text{hp} \in [-50,1.352]$;  & When the hypothesis is difficult enough to read, \\ 
    $\mathrm{PMI}(\text{label}_{\text{entailment}}, \text{flesch\_reading\_ease}_\text{hp}) >0.585$;
    & the sample tends to be labeled as entailment. \\
    \cmidrule(lr){1-2}
    
    $\text{male}_\text{hp} >2$, $\mathrm{PMI}(\text{label}_{\text{neutral}}, \text{male}_\text{hp}) >0.317$;
    & \multirow{2}[2]{*}{\makecell{Hypotheses with gender bias words \\ (male/female) tend to be neutral.}}  \\
     $\text{female}_\text{hp} >2$, $\mathrm{PMI}(\text{label}_{\text{neutral}}, \text{male}_\text{hp}) >0.377$; & \\
    \cmidrule(lr){1-2}
    
     $X = \text{len}_\text{pm} - \text{len}_\text{hp}$, if $X \in [8,30]$,  & \multirow{3}[1]{*}{\makecell[l]{When the length difference of hypothesis and premise is small \\  enough ([0,7]), the sample tends to be entailment, and when \\   it is large enough ([8,30]) the sample tends to be entailment.}} \\
     $\mathrm{PMI}(\text{label}_{\text{entailment}}, \text{len}_\text{pm} - \text{len}_\text{hp}) >0.084$;     & \\
     while $X \in [0,7]$;  $\mathrm{PMI}(\text{label}_{\text{neutral}}, \text{len}_\text{pm} - \text{len}_\text{hp})=0.045$ & \\
     \cmidrule(lr){1-2}
    
    % $X = \text{len}_\text{pm} - \text{len}_\text{hp}$, if $X \in [8,30]$,  &When the length difference between hypothesis and premise is \\
    % $\mathrm{PMI}(\text{label}_{\text{entailment}}, \text{len}_\text{pm} - \text{len}_\text{hp}) >0.084$;     & small enough ([0,7]), the sample tends to be entailment, and  \\
    %  while $X \in [0,7]$;  $\mathrm{PMI}(\text{label}_{\text{neutral}}, \text{len}_\text{pm} - \text{len}_\text{hp})=0.045$ & when it is large enough ([8,30]) the sample tends to be entailment. \\
    % \cmidrule(lr){1-2}
    
    $X = \text{len}_\text{pm} + \text{len}_\text{hp}$, if $X  \in [4,13]$,   & \multirow{3}[1]{*}{\makecell[l]{When the sum of the lengths of hypothesis and premise \\ is small enough,  the sample tends to be entailment, and when \\ it is large enough it tends to be neutral.}} \\
    $\mathrm{PMI}(\text{label}_{\text{entailment}}, \text{len}_\text{pm} + \text{len}_\text{hp}) =0.259$; & \\
    if $X >22$, $\mathrm{PMI}(\text{label}_{\text{neutral}}, \text{len}_\text{pm} + \text{len}_\text{hp}) >0.105$; &\\
    \cmidrule(lr){1-2}
   
    $X = \text{len}_\text{pm} / \text{len}_\text{hp}$, if $X < 2$, & \multirow{3}[1]{*}{\makecell[l]{When the lengths of hypothesis and premise are close enough, \\ the samples tend to be neutral, and when their lengths are \\ sufficiently different, samples tend to be entailment.}} \\
    $\mathrm{PMI}(\text{label}_{\text{neutral}},\text{len}_\text{pm} / \text{len}_\text{hp}) >0.094$; & \\
    if $X > 2$,
    $\mathrm{PMI}(\text{label}_{\text{entailment}},\text{len}_\text{pm} / \text{len}_\text{hp}) >0.141$; & \\
    \cmidrule(lr){1-2}
    
     $\mathrm{PMI}(\text{label}_{*},\text{len}_\text{pm}) \approx 0$; & \multirow{2}[1]{*}{\makecell[l]{The length and gender features of the premise are \\ irrelevance  with the label. }}\\
     & \\
     
    % $\text{pmi(premise\_len, label)} \approx 0$ & \multicolumn{1}{p{17.5em}}{The length and gender features of the premise are irrelevance  with the label. } \\
    \midrule
    \textbf{SST2} &  \\
    \midrule
    $\text{len}_\text{sent} <7$, $\mathrm{PMI}(\text{label}_\text{positive},\text{len}_\text{sent}) = 0.06$ & \multirow{2}[2]{*}{\makecell[l]{Sentences that are long enough tend to be negative, \\ while sentences that are short enough tend to be positive.}} \\
    $\text{len}_\text{sent}  >7$,  $\mathrm{PMI}(\text{label}_\text{negative},\text{len}_\text{sent}) > 0$ \\
    \cmidrule(lr){1-2}
    $\text{female}_\text{sent}  \in [4.8,5.4]$, $\mathrm{PMI}(\text{label}_\text{positive},\text{female}_\text{sent}) =0.58$ & \multirow{4}[2]{*}{\makecell[l]{Sentences with low female bias tend to be negative, \\with high female bias tend to be positive; \\ while sentences with high male bias tend to be negative.}} \\
    
    $\text{female}_\text{sent}<0.6$, $\mathrm{PMI}(\text{label}_\text{negative},\text{female}_\text{sent}) =0.021$ & \\
    
     $\text{male}_\text{sent}<1.2$, $\mathrm{PMI}(\text{label}_\text{positive},\text{male}_\text{sent}) =0.018$ & \\
      $\text{male}_\text{sent}>1.2$, $\mathrm{PMI}(\text{label}_\text{negative},\text{male}_\text{sent}) >0.068$ & \\
    
    % sentence with female bias $<0.6$; $\text{pmi(entailment)}=0.021$ &  \\
    % sentence with male name $<1.2$; $\text{pmi(neutral)}=0.018$ &  \\
    % sentence with male name $>1.2$; $\text{pmi(entailment)} >0.068$ &  \\
    \bottomrule
    \end{tabular}%
    \caption{Observations and conclusions of bias analysis with PMI on the SNLI and GLUE-SST2 dataset. ``\textit{hp}'' and ``\textit{pm}'' denote \textit{hypothesis} and \textit{premise}, respectively. ``$\text{len}$'' is a function that computes the length of a sentence. ``\textit{sent}'' denotes ``\textit{sentence}''.}
  \label{tab:app-pmi-analysis}%
\end{table*}%
% \footnote{\url{http://3.23.213.76/#/normal_dataset/617794bfb7314cb4146d2384/dataset_bias}}

\subsection{Bias}
\label{app:bias}
\paragraph{PMI for Sentiment Classification}
Taking the sentiment classification task as an example, we can use PMI to detect whether sentence length can indicate sentiment polarity. Given a sentence length sequence $L = \{l_1, l_2, \cdots, l_n \}$ with $n$ sentences, and a category sequence $C=\{c_1, c_2, \cdots, c_m\}$ with $m$ categories, the correlation measure PMI between sentence length and category can be defined as:
    \begin{equation}
        \phi_{\mathrm{pmi}}(c_i,l_j) = \log(\frac{p(c_i,l_j)}{p(c_i)p(l_j)}),
    \end{equation}
     where $c_i$ and $l_j$ denote the sentence length of the $i$-th sentence and the $j$-th category, respectively.

\paragraph{Gender Bias}
Given a male dictionary $K_{\text{male}}=[w_{m,1}, w_{m,2}, \dots, w_{m,k_1}]$ with $k_1$ words, female dictionary $K_{\text{female}}=[w_{f,1}, w_{f,2}, \dots, w_{f,k_2}]$ with $k_2$ words, and a dataset $D=[s_1, s_2, \dots, s_N]$ with $N$ samples, the gender bias $gb$ of dataset $D$ can be defined as:
\begin{align}
    b_m &= N_{\text{male}} / N, \\
    b_f &= N_{\text{female}} / N, \\
    gb &= b_m / b_f, 
\end{align}
where $b_m$ and $b_f$ is the degree to which the dataset is biased towards men and towards women, respectively.
$N_\text{male}$ and $N_{\text{female}}$ represent the number of words in the dataset D that appear in the dictionary $K_{\text{male}}$ and the number of words in the dictionary $K_{\text{female}}$, respectively. N is the sample size of dataset D. 

% \subsection{Additional Case study}
% \label{app:additional_case_study}
% \begin{figure}
%     \centering
%     \includegraphics[width=0.6\linewidth]{fig/case/gender.png}
%     \caption{Gender bias analysis on SNLI.}
%     \label{fig:gender_bias}
% \end{figure}

% \paragraph{Gender Bias}
% We also showcase the gender bias analysis on \texttt{SNLI} as illustrate in Fig.~\ref{fig:gender_bias}.
% %Our bias analysis is based on the proportion of male words (female words) contained in the sample to the total number of samples.
% We can find that the samples in the  \texttt{SNLI} dataset contain more male-oriented words than females ($\text{male}(0.62)>\text{female}(0.38)$).

\subsection{Calculation for Language Map}
In language map, each country will be assigned a number that can be obtained by following steps:
(1) for each country, collect the information that the languages spoken in this country and the proportion of people speaking each language.
(2) for each data set, record the language of the data set
(3) for each language, count the number of data set that belong to the language
(4) for each language in the country, multiply the ration of the language and the number of data set belong to the language. Finally sum the score of all languages in the country.

\subsection{Customized Operation}
\label{app:customized_operations}
\begin{lstlisting}
from datalabs import load_dataset
from featurize import featurize

# Operation definition
@datalabs.feature
def get_length(text):
    return len(text.split(" "))
    

# Load dataset    
dataset = load_dataset("ag_news")["train"]   
# Apply operation
res = dataset.apply(get_length)
\end{lstlisting}

\subsection{Technical Implementation of Data Search}
\label{app:dataset_finder}
Our dataset search tool is designed to take as input a natural language description of a method and compare it against a search corpus of datasets.

We train our retrieval model with the Tevatron package.\footnote{\url{https://github.com/texttron/tevatron}} The retrieval algorithm we use is effectively identical to Dense Passage Retrieval (DPR, \citet{karpukhin-etal-2020-dense}).
Under this dual-encoder framework, the search corpus is indexed by encoding each document using the \verb+CLS+ embedding from BERT \citep{DBLP:conf/naacl/DevlinCLT19}. When our system receives a query, we first compute its embedding (again using the \verb+CLS+ embedding from BERT), then we rank the top documents using approximate nearest neighbor search \citep{faiss} on the shared inner product space of embeddings:
$$
    \text{score}(q, d) = \verb+CLS+(\text{BERT}(q))^T \verb+CLS+(\text{BERT}(d))
$$
As a supervised learning-based retrieval method, this approach requires a large training set. To effectively generate a large training set, we adopt an automatic method for constructing annotations. We make the key observation that published AI/ML research papers reveal both a system description (contained in the abstract) as well as the datasets used to train or evaluate the system (usually found in the ``Results'' or ``Experiments'' section). 

We use the abstracts of real papers as a proxy for natural language method descriptions, but we do not expect users to submit abstract-length queries into our system. Therefore, we pass these abstracts through the ``TLDR'' scientific abstract summarization system \citep{cachola-etal-2020-tldr} to generate brief method descriptions.

We next automatically extract the datasets used by a given paper, which are used as a proxy for the relevant (positive) documents for each query during training. We extract these using a heuristic: for a given paper, if it mentions a dataset by name twice in the ``Results``, ``Experiments``, or ``Methods`` section and also cites the paper that introduces the dataset, we register this dataset as being used by the given paper. By manually inspecting 200 automatic dataset tags, we found over 90\% of the tags from this method were correct.

We also support traditional keyword queries in our system. To support these queries, we duplicate each example in our training set to replace the natural language description ``query'' with a keyword query. To generate keyword queries, we pass the abstract through a keyphrase extraction system trained on OpenKP \citep{Xiong2019OpenDW}. We then train a single retriever using a training set containing these two heterogenous types of queries.

\end{document}